\documentclass[10pt,twocolumn,letterpaper]{article}

\usepackage[pagenumbers]{cvpr} 

\usepackage{graphicx}
\usepackage{amsmath}
\usepackage{amssymb}
\usepackage{booktabs}
\usepackage{algorithm, algorithmic}

\usepackage{xcolor}         
\usepackage{pifont}

\definecolor{myred}{RGB}{220,50,47} 
\definecolor{mygreen}{RGB}{133,153,0}
\definecolor{commentcolor}{RGB}{133,153,0}

\newcommand*{\affmark}[1][*]{\textsuperscript{#1}}
\definecolor{urlcolor}{rgb}{0.93,0.01,0.55}

\usepackage[pagebackref,breaklinks,colorlinks]{hyperref}
\usepackage{enumitem}
\usepackage{color,colortbl}
\usepackage{tocloft}
\usepackage{titletoc}

\usepackage{amsthm}
\newtheorem{theorem}{Theorem}[section]

\newtheorem{remark}[theorem]{Remark}

\usepackage[capitalize]{cleveref}

\DeclareMathOperator*{\argmin}{arg\!\min}  

\def\cvprPaperID{****} 
\def\confName{arXiv}
\def\confYear{2023}

\begin{document}

\title{Improving Negative-Prompt Inversion via Proximal Guidance}
\title{ProxEdit: Improving Tuning-Free Real Image Editing with Proximal Guidance}
\title{Improving Tuning-Free Real Image Editing with Proximal Guidance}

\author{Ligong Han\affmark[1]\quad Song Wen\affmark[1]\quad Qi Chen\affmark[2]\quad Zhixing Zhang\affmark[1]\quad Kunpeng Song\affmark[1]\quad Mengwei Ren\affmark[3]\\Ruijiang Gao\affmark[4]\quad Anastasis Stathopoulos\affmark[4]\quad Xiaoxiao He\affmark[1]\quad Yuxiao Chen\affmark[1]\quad Di Liu\affmark[1]\\Qilong Zhangli\affmark[1]\quad Jindong Jiang\affmark[1]\quad Zhaoyang Xia\affmark[1]\quad Akash Srivastava\affmark[5]\quad Dimitris Metaxas\affmark[1]\\
{\affmark[1]Rutgers University\quad\affmark[2]Laval University\quad\affmark[3]New York University\quad\affmark[4]UT Austin\quad\affmark[5]MIT-IBM AI Lab}
}

\maketitle

\begin{abstract}
DDIM inversion has revealed the remarkable potential of real image editing within diffusion-based methods. However, the accuracy of DDIM reconstruction degrades as larger classifier-free guidance (CFG) scales being used for enhanced editing. Null-text inversion (NTI) optimizes null embeddings to align the reconstruction and inversion trajectories with larger CFG scales, enabling real image editing with cross-attention control. Negative-prompt inversion (NPI) further offers a training-free closed-form solution of NTI. However, it may introduce artifacts and is still constrained by DDIM reconstruction quality. To overcome these limitations, we propose proximal guidance and incorporate it to NPI with cross-attention control. We enhance NPI with a regularization term and inversion guidance, which reduces artifacts while capitalizing on its training-free nature. Additionally, we extend the concepts to incorporate mutual self-attention control, enabling geometry and layout alterations in the editing process. Our method provides an efficient and straightforward approach, effectively addressing real image editing tasks with minimal computational overhead\footnote{Code:~\href{https://github.com/phymhan/prompt-to-prompt}{\color{urlcolor}{https://github.com/phymhan/prompt-to-prompt}}.}.
\end{abstract}

\section{Introduction}
\label{sec:intro}
\begin{figure}[t]
  \centering
  \includegraphics[width=1\linewidth]{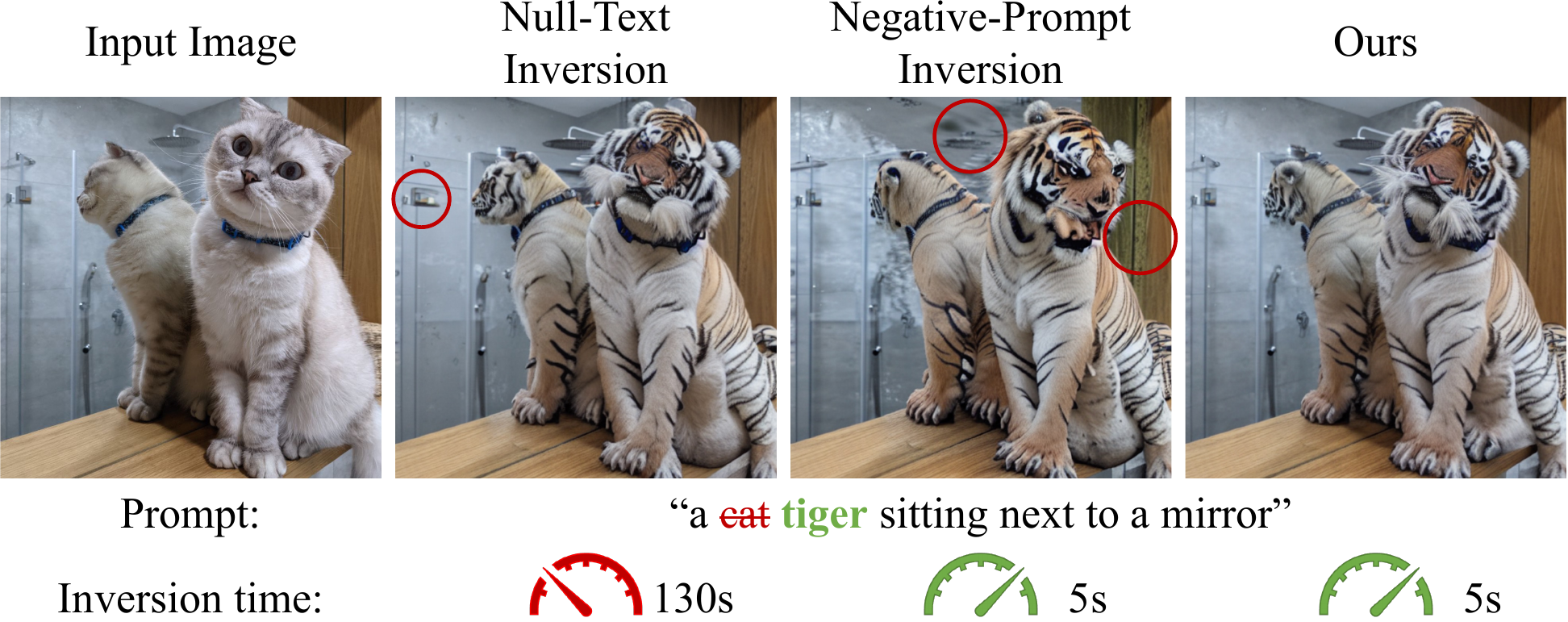}
\caption{\textbf{Proximal Negative-Prompt Inversion.} A comparison of editing quality between Null-text inversion (NTI), Negative-prompt inversion (NPI), and our proposed method (ProxNPI). The bottom row represents the time required for inversion. Our approach incorporates the fast inversion capability of NPI without the need for test-time optimization, thereby incurring only minimal additional cost during inference.}
  \label{fig:teaser}
\end{figure}
\begin{figure}[t]
  \centering
  \includegraphics[width=1\linewidth]{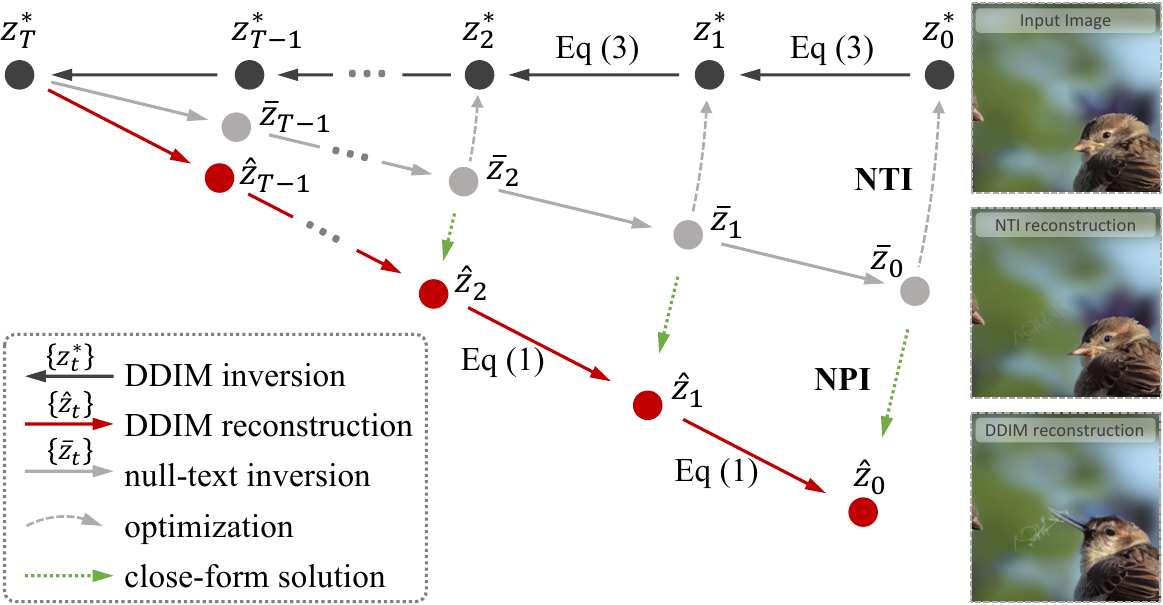}
\caption{Negative-Prompt Inversion (``NPI'') is the exact {\em closed-form} solution if we solve Null-text inversion (``NTI'') on the DDIM reconstruction sequence $\{\hat z_t\}$.}
  \label{fig:inversion}
\end{figure}
Diffusion-based methods have emerged as popular approaches for real image editing, with many of these methods utilizing DDIM inversion (a deterministic inversion method proposed in Denoising Diffusion Implicit Models~\cite{song2021denoising}). DDIM inversion is known to yield accurate reconstructions when using null embeddings or source prompts with a classifier-free guidance~\cite{ho2021classifier} (CFG) scale of 1. However, in order to achieve better editing capabilities, it is often necessary to use a CFG scale significantly larger than 1. Unfortunately, this scaling can lead to inaccurate reconstructions of the source image, which hinders the editing quality. This phenomenon is also observed in prompt-to-prompt~\cite{hertz2022prompt} editing scenarios.
\begin{figure*}[t]
  \centering
  \includegraphics[width=0.95\linewidth]{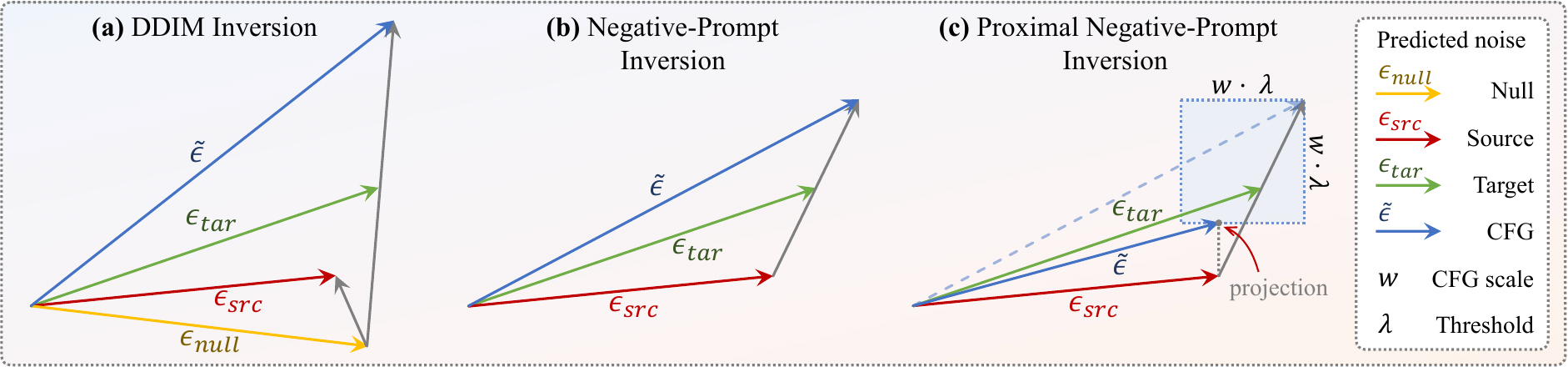}
\caption{Illustration of a single inference step using classifier-free guidance (CFG) with a scale $w=2$. All methods initially utilize DDIM inversion~\cite{song2021denoising} with the source prompt (and $w=1$). During the inference process: (a) direct sampling is performed using the target prompt; (b) the null embedding is replaced with the source prompt embedding; (c) a proximal gradient step is applied to the scaled noise difference $(\epsilon_{tar}-\epsilon_{src})$ following step (b). Here, we are visualizing soft-thresholding with a threshold $\lambda$, which corresponds to L1 regularization on $\tilde{\epsilon}$. If all values are clamped to zero, resulting in ProxNPI reducing to DDIM reconstruction. Conversely, when all values are retained after thresholding, ProxNPI reduces to NPI. }
  \label{fig:method}
\end{figure*}
To address this limitation, Null-text inversion~\cite{mokady2022null} (NTI) was introduced as a solution. NTI employs pivotal inversion by optimizing the null embedding(s), ensuring that the reconstruction trajectory aligns with the inversion trajectory even under a larger CFG scale. While NTI has a lightweight parameter set, it requires per-image optimization, which can be time-consuming. To eliminate the need for optimization in NTI, Negative-prompt inversion~\cite{miyake2023negative} (NPI) offers a closed-form solution. By assuming equal predicted noises between consecutive timesteps of the diffusion model, NPI elegantly demonstrated that the solver of NTI is equivalent to the source prompt embedding. However, NPI may occasionally introduce artifacts due to its underlying assumptions.

Building upon the remarkable results of NPI, we enhance it by incorporating a regularization term to improve the reconstruction of the source image. Moreover, we recognize that NPI is still constrained by the reconstruction quality of DDIM inversion, unable to correct errors introduced during the reconstruction process. To overcome this, we introduce a inversion guidance technique that performs one-step gradient descent on the current latent, aligning it with the inversion latents. The resulting algorithm offers a straightforward approach with negligible computational overhead.

Furthermore, as NTI and NPI are primarily designed for Cross-Attention Control~\cite{hertz2022prompt}, which focuses on texture and appearance changes, we extend our method to integrate proximal guidance into the Mutual Self-Attention Control framework~\cite{cao2023masactrl}. This integration allows for geometry and layout alterations in real image editing tasks.
In summary, our proposed method combines the benefits of NPI, inversion guidance, and a regularization term to provide an effective and efficient optimization-free solution for real image editing. We demonstrate its applications in NPI with Cross-Attention Control and Mutual Self-Attention Control, showcasing its versatility and potential impact.

\section{Related Work}
\label{sec:related}
Image generation with text guidance has been well-explored in image synthesis field~\cite{zhu2019dm, tao2020df, xu2018attngan,han2020robust,han2021dual,han2020unbiased,ren2022image,zhang2021cross,ye2021improving, ramesh2021zero, han2022ae,yu2022scaling, chen2023revisiting, hiclip, oh2001image, abdal2021styleflow, harkonen2020ganspace, patashnik2021styleclip,abdal2022clip2stylegan,han2022show,han2023svdiff}.
Recent development of text-to-image (T2I) diffusion models~\cite{sohl2015deep,song2019generative,ho2020denoising,song2020score,nichol2021improved,song2021denoising,gu2022vector,song2023consistency,chang2023muse} introduced new solutions to this task.
In particular, T2I diffusion models trained with large-scale image-caption pairs have shown impressive generation ability~\cite{ramesh2022hierarchical, saharia2022photorealistic, nichol2021glide, rombach2022high}.
The development of large-scale T2I models provides a giant and flexible design space for image manipulation methods leveraging the pre-trained model.
Recent works propose novel controlling mechanisms tailored for these T2I models~\cite{zhan2021multimodal, jiang2023avatarcraft, wang2023patch, wu2023sin3dm,zhang2023text,zhang2023robustness, lee2023soundini,li2023snapfusion,li2023layerdiffusion, chefer2023hidden,dutt2023parameter,avrahami2023break, huang2023zero,wang2023compositional}.

\noindent \textbf{Diffusion-based image editing.}
Many recent works fine-tune the pre-trained T2I models with a few personalized images to keep the context information~\cite{kumari2022multi,li2023stylediffusion,orgad2023editing,shi2023instantbooth,chen2023disenbooth,zhang2023prospect,sohn2023styledrop}.
Wide design choices have been explored in this direction.
Textual-Inversion~\cite{gal2022image,daras2022multiresolution,voynov2023p+}-based methods propose fine-tuning the text embedding.
Dreambooth~\cite{ruiz2022dreambooth} fine-tunes the whole model.
~\cite{kumari2022multi} fine-tunes the cross-attention layers in the UNet of Stable-Diffusion model.
These methods require hundreds of iterations at the fine-tuning stage to capture the identity information.
For better efficiency, more techniques~\cite{hu2021lora,han2023svdiff,mou2023t2i,li2023gligen} are developed by reducing the number of parameters optimized at fine-tuning stage.
While fine-tuning the pre-trained T2I model shows extraordinary results, the test-time efficiency of these methods remains a great challenge.
SEGA~\cite{brack2023sega} discovers that target concept can be encoded using latent dimensions falling into the upper and lower tail of the distribution. 

\noindent \textbf{Inversion-based image editing.}
DDIM inversion~\cite{song2021denoising} is widely adopted in editing tasks by deterministically encoding the original image into a latent noise that can be accurately reconstructed. Leveraging pivotal inversion on null-text embeddings, Null-text Inversion~\cite{mokady2022null} improved the identity preservation of the edit. However, all these methods rely on optimization at test-time for accurate reconstruction. Negative-prompt inversion (NPI)~\cite{miyake2023negative} further avoided the computation cost for the optimization while achieves competitive results as null-text inversion.
One can also interpret NPI as performing Delta Denoising Score (DDS)~\cite{hertz2023delta} on the same noisy image.
\begin{algorithm}[t]
	\renewcommand{\algorithmicrequire}{\textbf{Input:}}
	\renewcommand{\algorithmicensure}{\textbf{Output:}}
	\caption{Proximal Negative-Prompt Inversion}
	\label{alg:1}
	\begin{algorithmic}[1]
		\REQUIRE Given source original sample $z_0$, source condition $C$, target condition $C'$, denoising model $\epsilon_{\theta}$, proximal function $\text{prox}_{\lambda}(\cdot)$.
        \STATE $\bar {z}_T=\text{DDIMInvert}(z_0, C, w=1)$
            \STATE $\tilde z_T = {\bar {z}_T}$
		\FOR{$t = T$ to $1$}
            \STATE $\tilde\epsilon_{src} = \epsilon_{\theta}(\tilde z_t, t, C)$
            \STATE $\tilde\epsilon_{tar} = \epsilon_{\theta}(\tilde z_t, t, C')$
            \STATE $\tilde\epsilon = {\tilde\epsilon}_{src} + w \cdot \text{prox}_{\lambda}(\tilde\epsilon_{tar}- \tilde\epsilon_{src})$
            \STATE $M = |\tilde\epsilon_{tar}- \tilde\epsilon_{src}| \leq \lambda$
            \STATE $\tilde z_0 = \frac{1}{\sqrt{\alpha_t}}\tilde z_t - \sqrt{\frac{1}{\alpha_t}-1}\tilde\epsilon$
            \STATE $\tilde z_{t-1} = \sqrt{\alpha_{t-1}}\tilde z_0+\sqrt{1-\alpha_{t-1}}\tilde\epsilon$
            \IF {inversion guidance \AND $t<T_{inv}$}
                \STATE $\tilde z_{t-1} = \tilde z_{t-1} - \eta M\odot(\tilde z_{t-1} - z^*_{t-1})$
            \ENDIF
		\ENDFOR
		\STATE \textbf{return} $\tilde z_0$
	\end{algorithmic}  
\end{algorithm}
\begin{figure}[t]
  \centering
  \includegraphics[width=1\linewidth]{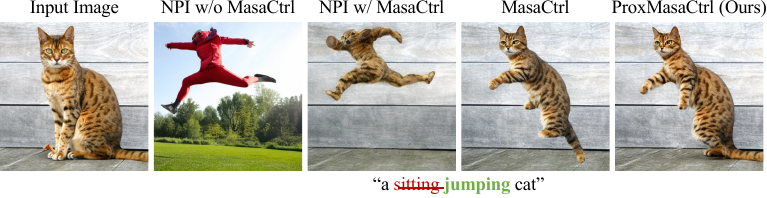}
\caption{\textbf{Applying Negative Prompt Inversion (NPI) to Mutual Self-Attention Control (MasaCtrl~\cite{cao2023masactrl}).} Directly applying NPI to MasaCtrl by substituting the null embedding with the source prompt embedding leads to the presence of strange artifacts (labeled as ``NPI w/ MasaCtrl''). In our approach, we solely replace the null embedding with the source prompt in the DDIM reconstruction branch.}
  \label{fig:masa_intro}
\end{figure}
\begin{figure*}[t]
  \centering
  \includegraphics[width=1\linewidth]{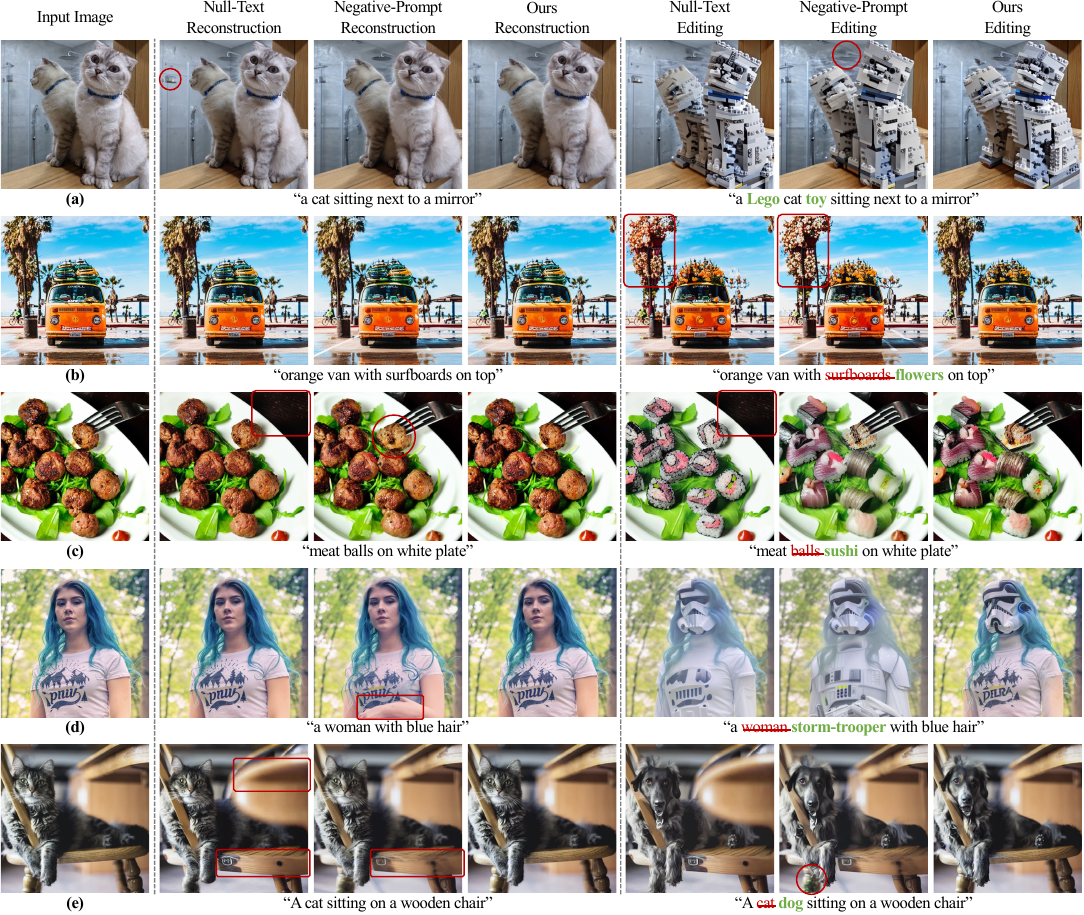}
\caption{\textbf{Qualitative comparisons of inversion methods.} The figure showcases qualitative comparisons among Null-text inversion (NTI)~\cite{mokady2022null}, Negative-prompt inversion (NPI)~\cite{miyake2023negative}, and our proposed method (ProxNPI). Each row demonstrates the reconstruction results (columns 2-4) and editing results (columns 5-7) for the respective methods. Inversion guidance is employed to address minor errors in DDIM reconstruction. Errors or artifacts are marked using red circles or boxes. The comparisons highlight instances where NPI fails to retain specific image details (a), both NTI and NPI introduce undesired changes (b), the inversion guidance aids in recovering missing details (c), our method exhibits better background preservation (d), and NTI/NPI exhibit reconstruction errors (e).}
  \label{fig:compare}
\end{figure*}

\begin{figure*}[t]
  \centering
  \includegraphics[width=1\linewidth]{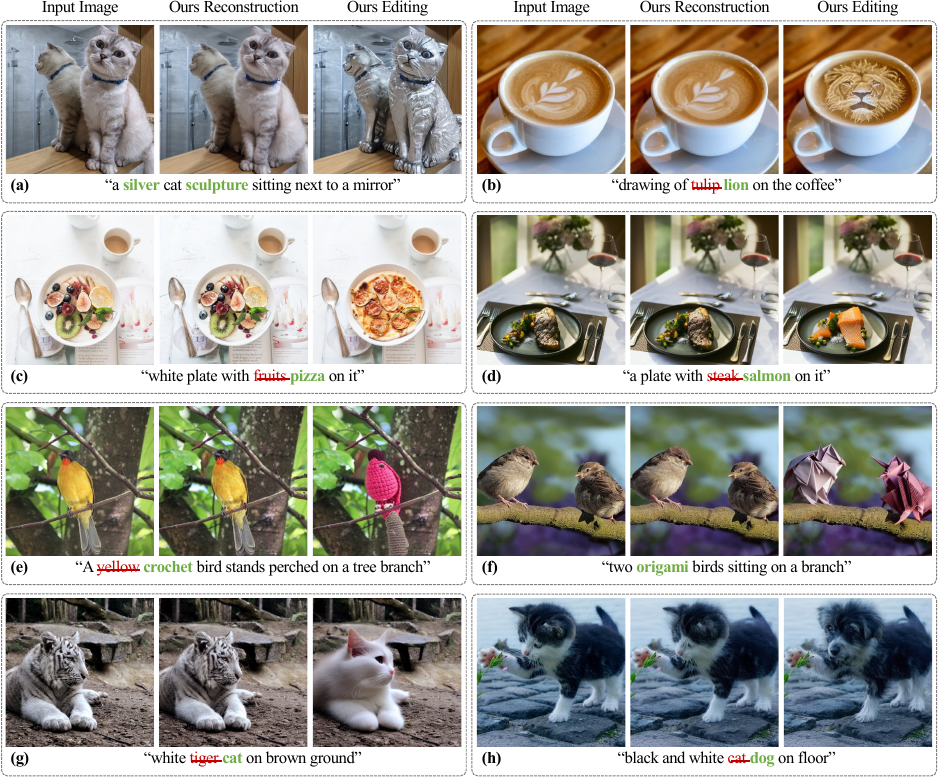}
\caption{\textbf{Additional visual editing results.} More visual editing results for our method are presented, along with their corresponding prompts. Inversion guidance is applied for examples (c), (e), (f), and (g) due to imperfect DDIM reconstructions in these cases.}
  \label{fig:result}
\end{figure*}

\begin{figure*}[t]
  \centering
  \includegraphics[width=1\linewidth]{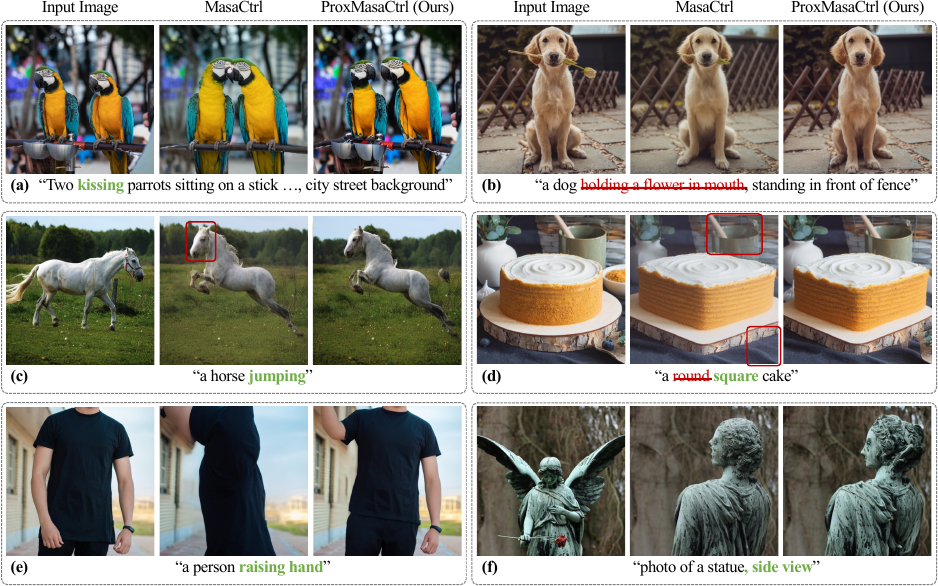}
\caption{\textbf{Enhancing MasaCtrl~\cite{cao2023masactrl} with Proximal Guidance.} Our proposed proximal guidance offers improvements in cases where MasaCtrl exhibits instability or introduces undesired changes that users wish to retain. MasaCtrl may introduce slight color shifts in the main subject(s) or background, as depicted in (a), (b), (c), and (f). However, with proximal guidance, the background and intended details are better preserved. For instance, in (a), the two steel bowls at the bottom and the person holding a phone at the right edge; in (b), the fence on the upper right and the dog's mouth; in (c), the reins on the horse's head; and in (d), the cup and vase in the background.}
  \label{fig:masa_ours}
\end{figure*}

\begin{figure}[t]
  \centering
  \includegraphics[width=1\linewidth]{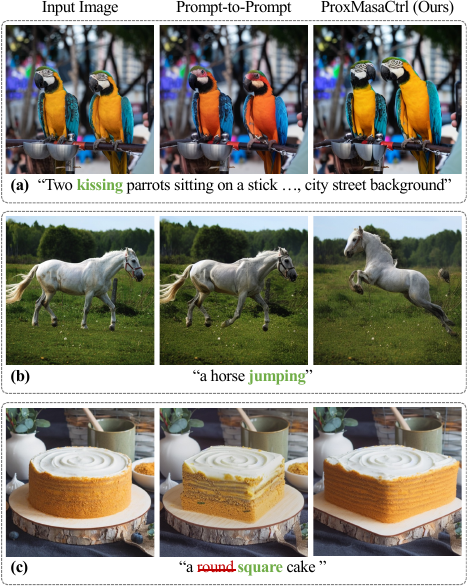}
\caption{\textbf{Qualitative Comparison of Prompt-to-Prompt~\cite{hertz2022prompt} and MasaCtrl~\cite{cao2023masactrl}.} Column 2 presents the editing results obtained using NPI + Cross-Attention Control (``Prompt-to-Prompt''), while column 3 displays the editing results of MasaCtrl with proximal guidance (``ProxMasaCtrl''). As anticipated, Prompt-to-Prompt, which is not specifically designed for shape or geometry changes, exhibits limited editing capabilities, as demonstrated in (a) and (b). In (c), Prompt-to-Prompt introduces a new texture on the cake that is absent in the input image, while (Prox-)MasaCtrl ``recycles'' the original texture.}
  \label{fig:masa_ptp}
\end{figure}

\begin{figure}[t]
  \centering
  \includegraphics[width=1\linewidth]{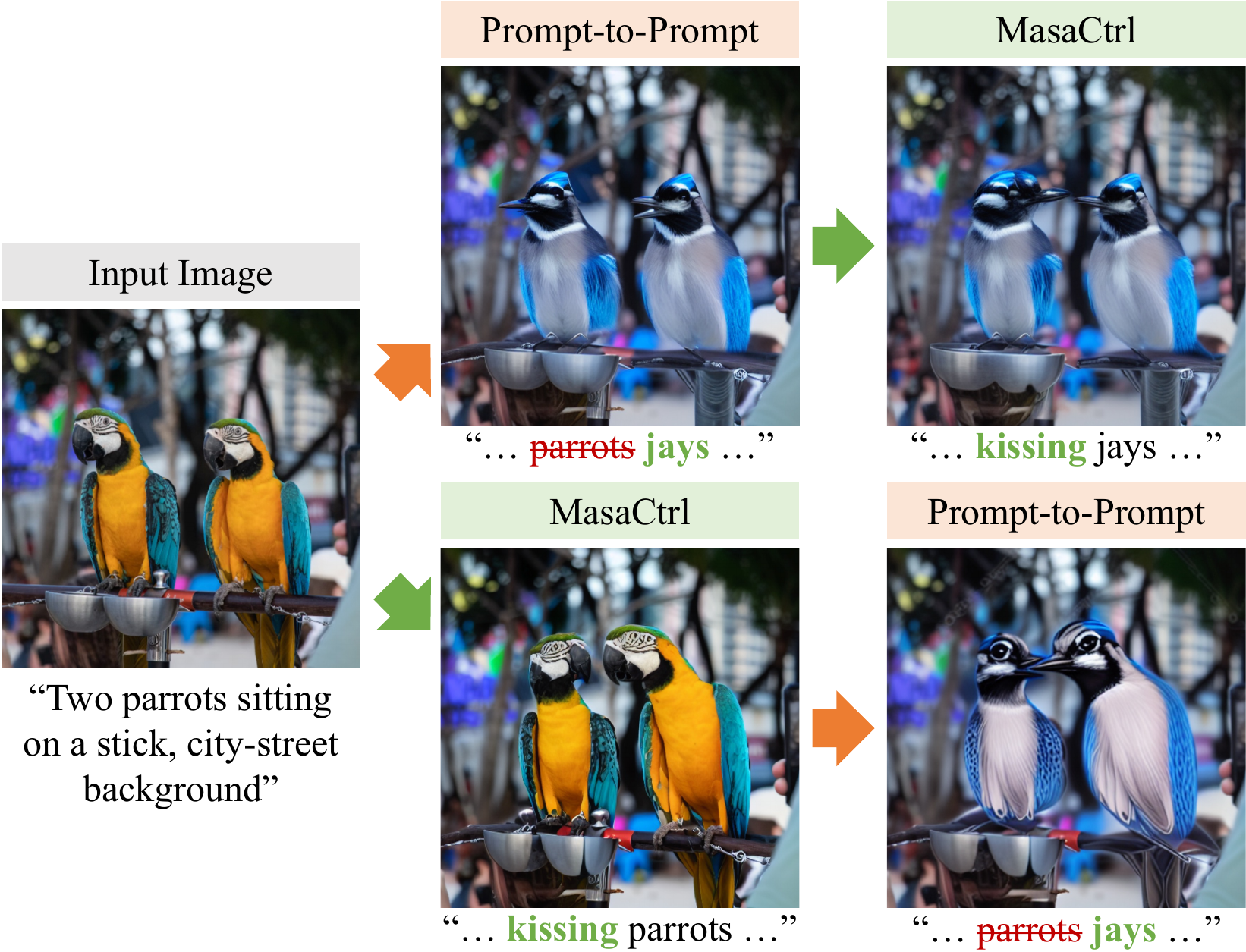}
\caption{\textbf{Editing texture and pose.} Sequentially applying Prompt-to-Prompt and MasaCtrl (first row) or in the reverse order (second row) to edit both texture and pose.}
  \label{fig:parrot_full}
\end{figure}

\begin{figure*}[t]
  \centering
  \includegraphics[width=0.8\linewidth]{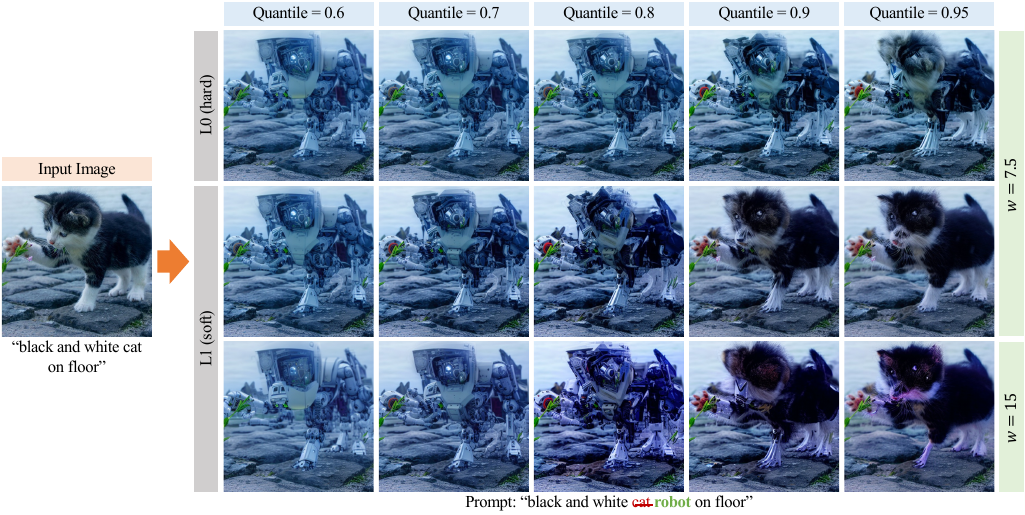}
\caption{\textbf{Ablation study of thresholds.} The figure shows visual results obtained by varying the threshold $\lambda$ from the 60\% to 95\% quantiles of the absolute noise difference. The first row represents the impact of hard-thresholding ($L0$), while the second and third rows show the effects of soft-thresholding ($L1$). Soft-thresholding induces less noticeable changes in the edited images compared to hard-thresholding, aligning with our expectations. Alternatively, increasing the CFG scale, such as with $w=15$, can enhance the prominence of the target attribute, although it may introduce an intensified contrast ratio and shifted color tone.}
  \label{fig:ablate}
\end{figure*}
\begin{figure}[t]
  \centering
  \includegraphics[width=1\linewidth]{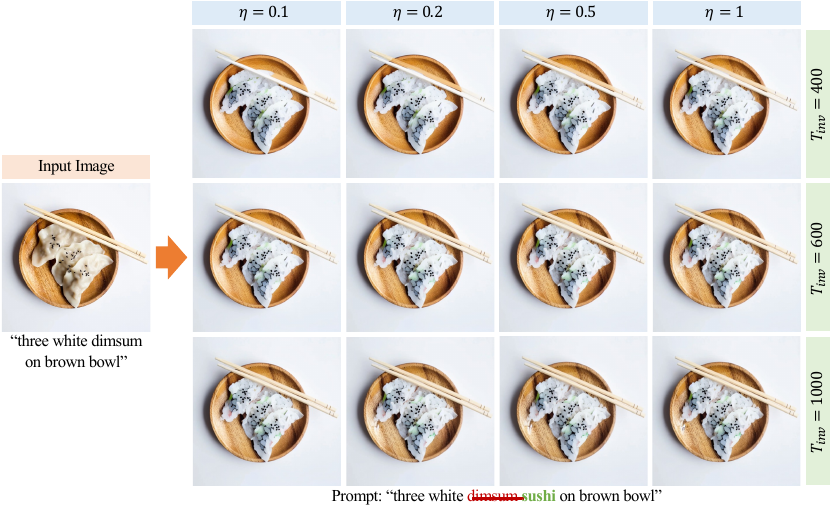}
\caption{\textbf{Ablation study of inversion guidance.} The figure shows visual results obtained by varying the stepsize of performing inversion guidance $\eta$ from the 0.1 to 1. The threshold is set to the 70\% quantile and hard-thresholding is used.}
  \label{fig:ablate_inv}
\end{figure}
\begin{figure*}[t]
  \centering
  \includegraphics[width=1\linewidth]{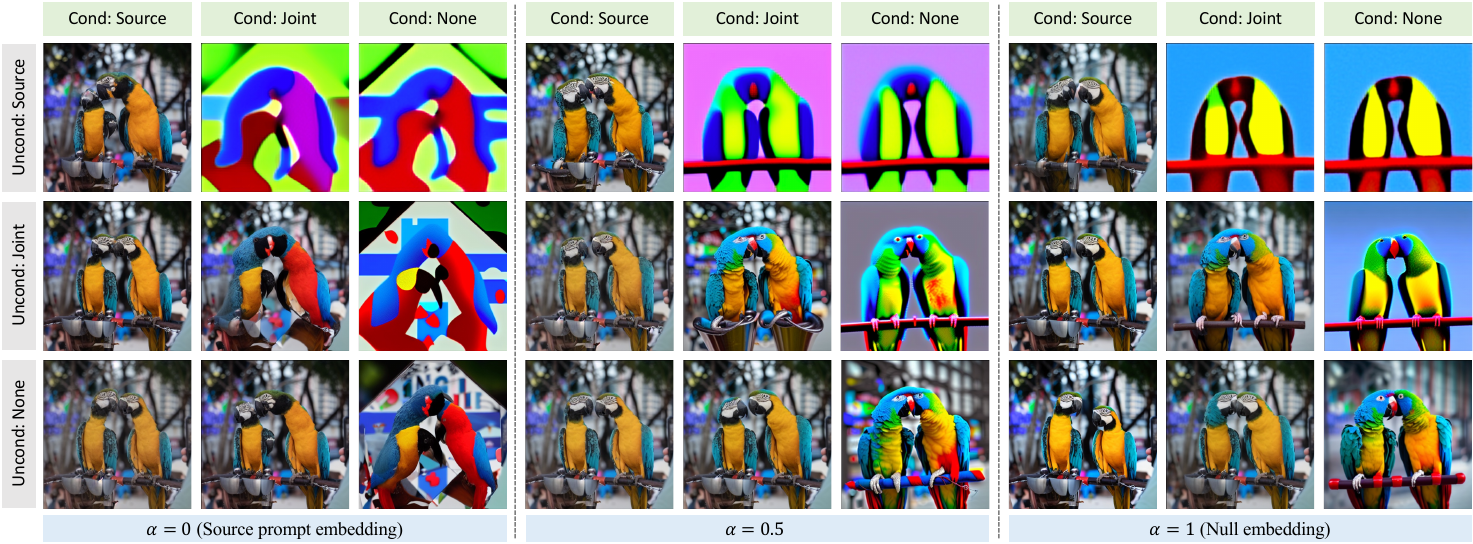}
\caption{\textbf{MasaCtrl feature injection ablation.} We ablate different feature injection strategies in MasaCtrl by varying the $\alpha$ parameter (in the synthesis branch) and querying different feature sets (``source'', ``joint'', ``none''). For all $\alpha$ we use $C_{src}$ in reconstruction branch. In the above example, we find that using ``joint'' or ``none'' for the unconditional noise improves results.}
  \label{fig:ablate_prox_masa}
\end{figure*}

\section{Method}
\label{sec:method}
\subsection{Background}
\label{sec:method:background}
\noindent \textbf{DDIM inversion.}
DDIM is a widely used deterministic sampling (if chosen to be) of DDPM. While DDPM follows a stochastic differential equation (SDE) process, DDIM corresponds to its ordinary differential equation (ODE) counterpart. The reverse DDIM process can be written as
\begin{align}
    z_{t-1}=&\frac{\sqrt{\alpha_{t-1}}}{{\sqrt\alpha_{t}}}z_t + \label{eq:ddim_reverse}\\
    &\sqrt{\alpha_{t-1}} \left(\sqrt{\frac{1}{\alpha_{t-1}}-1} - \sqrt{\frac{1}{\alpha_{t}}-1}\right)\epsilon_{\theta}(z_t, t, C),\nonumber
\end{align}
\noindent where $C$ is the given conditioning. To invert the given image, the latent variables can be estimated by reversing the above discrete ODE sampling process. By rearranging \cref{eq:ddim_reverse}, we have
\begin{align}
    z_{t}=&\frac{\sqrt{\alpha_{t}}}{{\sqrt{\alpha_{t-1}}}}z_{t-1} + \label{eq:ddim_forward_exact}\\
    &\sqrt{\alpha_{t}} \left(\sqrt{\frac{1}{\alpha_{t}}-1} - \sqrt{\frac{1}{\alpha_{t-1}}-1}\right)\epsilon_{\theta}(z_{t}, t, C).\nonumber
\end{align}
\noindent Note that $z_t$ appears at both sides. A common technique is to approximate $\epsilon_{\theta}(z_{t}, t, C)$ with $\epsilon_{\theta}(z_{t-1}, t-1, C)$, such that the inversion process can be solved by Euler method. Then, denote the sequence of latent variables from $z_0$ via DDIM inversion as $\{z_t^*\}_{t=1}^T$, we have
\begin{align}
    z_{t}^*=&\frac{\sqrt{\alpha_{t}}}{{\sqrt{\alpha_{t-1}}}}z_{t-1}^* + \label{eq:ddim_forward}\\
    &\sqrt{\alpha_{t}} \left(\sqrt{\frac{1}{\alpha_{t}}-1} - \sqrt{\frac{1}{\alpha_{t-1}}-1}\right)\epsilon_{\theta}(z_{t-1}^*, t-1, C).\nonumber
\end{align}

\noindent \textbf{Null-text inversion.} Using the classifier-free guidance (CFG~\cite{ho2021classifier}), the noise is estimated by
\begin{align}
    \tilde{\epsilon}_\theta(z_t, t, C, \emptyset) = w \epsilon_\theta(z_t, t, C) + (1-w) \epsilon_\theta(z_t, t, \emptyset)
\end{align}
If $w > 1$, the accumulated error on DDIM inversion will affect reconstruction accuracy. To address the problem, null-text inversion~\cite{mokady2022null} (NTI) optimizes a set of per-timestep null-text embeddings $\{\emptyset_t\}$ to track the DDIM inversion trajectory even under a large $w$. It first computes $\{z_t^*\}_{t=1}^T$ using DDIM inversion with $w = 1$. Then, after initializing ${\bar{z}}_T=z_T^*$, null-text inversion solves $\emptyset_t$ by performing the following optimizations for $t=T,\ldots,1$:
\begin{align}
    \min_{\emptyset_t}||z_{t-1}(\bar{z_t}, \emptyset_t, C)-z_{t-1}^*||_2^2.
\end{align}

\noindent \textbf{Negative-prompt inversion} 
(NPI~\cite{miyake2023negative}) overcomes the limitation of per-image optimization in null-text inversion by providing a {\em closed-form} solution, $\emptyset_t=C$, with minimal approximation. NPI validates this solution through induction: if $\emptyset_t=C$ and $\bar{z}_t=z^*_t$ hold for timestep $t$, they also hold for timestep $t-1$, by assuming $\epsilon_{\theta}(z_{t}^*, t, C) \approx \epsilon_{\theta}(z_{t-1}^*, t-1, C)$. We can verify that with $\emptyset_t=C$, NPI reconstruction with $w>1$ recovers the DDIM reconstruction,
\begin{align}
    \tilde{\epsilon}_\theta(z_t, t, C, C) &= w \epsilon_\theta(z_t, t, C) + (1-w) \epsilon_\theta(z_t, t, C)\nonumber\\
    &=\epsilon_\theta(z_t, t, C).
\end{align}
\noindent In fact, as demonstrated in \cref{fig:inversion} and the subsequent remark, NPI provides an exact solution without any approximation when tracking the DDIM reconstruction trajectory instead of the inversion trajectory:
\begin{remark}\label{remark:1}
    Negative-prompt inversion is the {\em exact} closed-form solution if we solve null-text inversion optimizations to track the DDIM reconstruction trajectory $\{\hat{z}_t\}$.
\end{remark}

\subsection{Proximal Negative-Prompt Inversion}
Negative-prompt inversion provides an elegant closed-form solution for computing null-text inverted null-embeddings, $\emptyset_t=C$. This solution intuitively aligns with the DDIM reconstruction process. \cref{fig:method} illustrates a single inference step using classifier-free guidance (CFG) with a scale parameter $w=2$. Initially, all methods employ DDIM inversion~\cite{song2021denoising} with the source prompt (and $w=1$). In \cref{fig:method}(a), we depict a baseline approach where direct sampling is performed using the target prompt. \cref{fig:method}(b) demonstrates the inference step of negative-prompt inversion, where CFG amplifies the editing direction of $(\tilde\epsilon_{tar}-\tilde\epsilon_{src})$. Intuitively, when the target prompt is close to the source prompt, the inference trajectory for editing should closely resemble the DDIM reconstruction trajectory. In fact, when the target condition $C'=C$, negative-prompt inversion exactly recovers DDIM reconstruction. However, we observe that negative-prompt inversion occasionally over-amplifies the editing direction $(\tilde\epsilon_{tar}-\tilde\epsilon_{src})$. To address this, we propose the addition of an extra loss term that encourages the CFG noise $\tilde \epsilon$ to align with $\tilde\epsilon_{src}$.

To accomplish this, we draw inspiration from the proximal gradient method~\cite{tibshirani2019prox,fessler2021prox} and introduce a regularization term to constrain $(\tilde\epsilon_{tar}-\tilde\epsilon_{src})$. This regularization is achieved through the use of a proximal function,
\begin{align}
    \text{prox}_{\lambda,L_p}(x) = \argmin_{z}{\frac{1}{2}\|z-x\|_2^2 + \lambda \|z\|_p}.
\end{align}
\noindent which encourages desired properties in the editing process. When $p=1$ (corresponding to $L1$ regularization), the solver takes the form of a soft-thresholding function,
\begin{align}
    [\text{prox}_{\lambda,L_1}(x)]_i = [S_{\lambda}(x)]_i =  \left\{ \begin{array}{rcl} x_i-\lambda & \mbox{if} & x_i > \lambda \\
    0 & \mbox{if} & -\lambda\leq x_i \leq \lambda \\
    x_i+\lambda & \mbox{if} & x_i < -\lambda
\end{array}\right.
\end{align}
\noindent with $\left[\tiny{~}\cdot\tiny{~}\right]_i$ denoting the $i$-th component. if $p=0$ (representing $L0$ regularization), the solver takes the form of a hard-thresholding function,
\begin{align}
    [\text{prox}_{\lambda,L_0}(x)]_i = \left\{ \begin{array}{rl} x_i & \mbox{if} \quad |x_i| > \sqrt{2\lambda} \\
    0 & \mbox{otherwise}.
\end{array}\right.
\end{align}
\noindent Since the value range of $(\tilde\epsilon_{tar}-\tilde\epsilon_{src})$ does not follow a standard Gaussian distribution, we employ a dynamic threshold rather than a fixed one by selecting a quantile of the absolute values $|\tilde\epsilon_{tar}-\tilde\epsilon_{src}|$. It is worth noting that when using hard thresholding with quantile thresholds, this approach is similar to SEGA~\cite{brack2023sega}. \cref{fig:method}(c) provides a visualization of a 2-D case when soft-thresholding is employed. If all values are clamped to zero, our method reduces to DDIM reconstruction. Conversely, when all values are retained after thresholding, our method simplifies to negative-prompt inversion.

\noindent \textbf{Inversion guidance.}
As discussed previously, NPI is still upper-bounded by the quality of DDIM reconstruction. Even if $\tilde \epsilon$ converges to $\epsilon_{src}$, it cannot correct errors in cases where DDIM reconstruction is imperfect. On the other hand, NTI tracks the DDIM inversion trajectory and thus does not have such limitation. This motivates us to introduce a {\em inversion guidance} by performing a single step of gradient descent on the current latent $\tilde{z}_{t-1}$, aiming to align it with the inversion latent $z^*_{t-1}$. This gradient descent step is applied only to the ``unedited'' region identified by the mask $M = |\tilde\epsilon_{tar} - \tilde\epsilon_{src}| \leq \lambda$, where we reuse the notation $\lambda$ to represent the {\em threshold} value. By choosing a step size $\eta$, the update can be expressed as $\tilde z_{t-1} \leftarrow \tilde z_{t-1} - \eta M \odot (\tilde z_{t-1} - z^*_{t-1})$, where $\eta=1$ corresponds to a complete replacement. The complete algorithm is outlined in \cref{alg:1}.
The algorithm can be thought of as an ADMM~\cite{boyd2011distributed} type of method that solves NTI on the DDIM inversion trajectory:
\begin{align}
    \min_{\emptyset_t}||z_{t-1}(\bar{z_t}, \emptyset_t, C)-\hat{z}_{t-1}||_2^2 \quad \text{s.t.} \quad \hat{z}_{t-1}={z}^*_{t-1}
\end{align}
\noindent where the objective is solved by NPI (see Remark~\ref{remark:1}) and the constraint is enforced by inversion guidance.

\subsection{Proximal Mutual Self-Attention Control}
Both NTI and NPI are designed to be used with Cross-Attention Control (or Prompt-to-Prompt~\cite{hertz2022prompt}) for real image editing. While Cross-Attention Control primarily focuses on changing the texture or appearance of a subject, recent methods have explored self-attention controlling mechanisms for manipulating geometry or layout~\cite{tumanyan2023plug,qi2023fatezero,cao2023masactrl,zhang2023real}. MasaCtrl~\cite{cao2023masactrl} proposes a Mutual Self-Attention Control mechanism that queries image content from the source input image. In this section, we aim to integrate proximal guidance into the MasaCtrl framework.

Although NTI/NPI and MasaCtrl operate through different mechanisms, they share the same goal of preserving specific content from the source image. Initially, we observe that directly incorporating NPI with MasaCtrl by substituting the null embedding with the source prompt embedding can lead to artifacts. This occurs because, without any cross-attention or self-attention control, this is equivalent to setting the source prompt as negative prompt. As illustrated in an example in \cref{fig:masa_intro}, using the source prompt as the negative prompt (``NPI w/o MasaCtrl'') generates a jumping person unrelated to the source image. When combined with MasaCtrl (``NPI w/ MasaCtrl''), the model is compelled to query cat features to render the same jumping person. Therefore, we propose using NPI solely in the reconstruction branch while retaining the null embedding in the synthesis branch:
\begin{align}
    \hat\epsilon &= {\hat\epsilon}_{src} + 1 \cdot (\hat\epsilon_{src}- \hat\epsilon_{src}), &\text{[reconstruction]}\nonumber\\
    \tilde\epsilon &= {\tilde\epsilon}_{null} + w \cdot \text{prox}_{\lambda}(\tilde\epsilon_{tar}- \tilde\epsilon_{null}). &\text{[synthesis]}\label{eq:masa_prox}
\end{align}
\noindent Here, we introduce proximal guidance to the term $(\tilde\epsilon_{tar}- \tilde\epsilon_{null})$. During model forward passes, the MasaCtrl mechanism forces both ${\tilde\epsilon}_{null}$ and ${\tilde\epsilon}_{tar}$ to query features from ${\hat\epsilon}_{src}$ (in the original MasaCtrl, the unconditional part ${\tilde\epsilon}_{null}$ queries from ${\hat\epsilon}_{null}$ in the reconstruction branch). By setting $\lambda$ to the 100\% quantile, $\tilde\epsilon$ converges to ${\tilde\epsilon}_{null} \approx {\hat\epsilon}_{src}$, degrading to a DDIM reconstruction. Hence, similar to ProxNPI, the introduced proximal guidance here also controls the proximity of the synthesized image to the source image.

\section{Experiment}
\label{sec:exp}
\subsection{Cross-Attention Control}
In this section, we present qualitative comparisons among Null-text inversion (NTI)~\cite{mokady2022null}, Negative-prompt inversion (NPI)~\cite{miyake2023negative}, and our proposed method (ProxNPI), as illustrated in \cref{fig:compare}. Each row in the figure showcases the reconstruction results (columns 2-4) and editing results (columns 5-7) for each method. It is worth noting that NPI reconstruction is equivalent to DDIM reconstruction~\cite{song2021denoising}, as discussed previously. For examples (c-e) in \cref{fig:compare}, we utilize inversion guidance since DDIM reconstruction still introduces minor errors. These errors or artifacts are highlighted using red circles or boxes.

In \cref{fig:compare}, we observe that NPI fails to retain the shower head in the mirror (a), while both NTI and NPI alter the leaves on the tree to flowers (b). In (c), both NTI and NPI exhibit imperfect reconstruction of the input image, whereas the missing detail is recovered through the inversion guidance incorporated in our method. Additionally, in (d), NPI introduces reconstruction errors, while our method demonstrates superior preservation of the background in edited images compared to NTI and NPI. Lastly, in (e), both NTI and NPI display reconstruction errors on the chair.

\noindent \textbf{Additional visual results.}
We present more visual editing results in \cref{fig:result}, along with the corresponding prompts provided underneath the images. Among the eight examples, inversion guidance is employed for examples (c), (e), (f), and (g) due to imperfections in DDIM reconstructions for these specific cases.

\subsection{Mutual Self-Attention Control}
We conducted qualitative comparisons between MasaCtrl~\cite{cao2023masactrl} and our proposed ProxMasaCtrl, as shown in \cref{fig:masa_ours}. Our ProxMasaCtrl incorporates proximal guidance to address the issues of instability and undesired changes occasionally observed with MasaCtrl. As shown, MasaCtrl can introduce slight color shifts in the main subject(s) and background, as demonstrated in examples (a), (b), (c), and (f). However, with proximal guidance, we achieve better preservation of the background and intended details. For instance, in example (a), the two steel bowls at the bottom and the person holding a phone at the right edge are preserved. Similarly, in example (b), the fence on the upper right and the dog's mouth are retained with improved fidelity. Additionally, in example (c), the reins on the horse's head are better maintained, and in example (d), the cup and vase in the background are better preserved.

\noindent \textbf{Comparing with Prompt-to-Prompt.}
As mentioned earlier, Prompt-to-Prompt~\cite{hertz2022prompt} is designed to alter the texture of a subject, whereas Mutual Self-Attention Control~\cite{cao2023masactrl} targets geometry and layout modifications. \cref{fig:masa_ptp} shows a visual comparison between these two attention controlling mechanisms for geometry editing. This comparison aims to illustrate their respective behaviors rather than establish the superiority of one over the other. Notably, in \cref{fig:masa_ptp}(c), Prompt-to-Prompt introduces a new texture on the cake, resembling its cross-section that deviates from the source image, while MasaCtrl preserves the original appearance. Additionally, Prompt-to-Prompt confines the rendered cake to a square shape within the cross-attention map of the original round cake, whereas MasaCtrl allows rendering outside the boundaries of the original cake. Note that the base of the cake is also changed to square.

\noindent \textbf{Simultaneous texture and geometry editing.}
We extend our approach by sequentially applying ProxNPI and ProxMasaCtrl to enable simultaneous editing of both texture and geometry, as shown in \cref{fig:parrot_full}. While this represents a preliminary exploration, we leave a more effective integration of these two controlling mechanisms for future research.

\subsection{Ablations}
\noindent \textbf{Thresholding.}
In \cref{fig:ablate} we present visual results obtained by setting the threshold $\lambda$ to the 60\%, 70\%, ..., 95\% quantiles of the absolute values of the noise difference. The first row illustrates the case of hard-thresholding (labeled as $L0$), while the second and third rows display the case of soft-thresholding (labeled as $L1$). As anticipated, we observe that soft-thresholding tends to introduce fewer changes to the edited images compared to hard-thresholding. Alternatively, we can increase the CFG scale, such as using $w=15$, to enhance the prominence of the target attribute. However, this approach may lead to an amplified contrast ratio and shifted color tone. We empirically find using hard-thresholding with quantile 0.7 usually gives good results.

\noindent \textbf{Inversion guidance.}
In \cref{fig:ablate_inv} we present visual results obtained by varying the stepsize $\eta$ for the inversion guidance, applied when $t<T_{inv}$. As observed, when the guidance strength $\eta$ and $T_{inv}$ are small, the reconstruction of chopsticks is incomplete, and the pattern on the bowl is missing. We empirically find that setting $T_{inv} \ge 600$ and $\eta \ge 0.2$ generally yields satisfactory results.

\noindent \textbf{MasaCtrl ablation.} In \cref{fig:ablate_prox_masa}, we conduct ablations of different mutual self-attention feature injection strategies in MasaCtrl. The synthesis branch utilizes the ``null embedding'' denoted as $C=\text{interp}(\alpha, C_{src}, C_{null})$, where the default setting uses the original null embedding with $\alpha=1$. We explore the visual effects by varying $\alpha$ and querying different feature sets, including ``source'' (the default strategy), ``joint'' (querying from both branches), and ``none'' (no feature injection). While $\alpha=1$ with ``source'' generally produces good results for both unconditional and conditional noises, we observe that using ``joint'' or ``none'' for the unconditional noise occasionally improves the outcomes.

\section{Discussion and Conclusion}
\label{sec:conclusion}
In this paper, we introduced proximal guidance, a versatile technique for enhancing diffusion-based tuning-free real image editing. We applied this technique to two concurrent frameworks: Negative-prompt inversion (NPI) and Mutual Self-Attention Control. The resulting algorithms, ProxNPI and ProxMasaCtrl, addressed limitations and achieved high-quality editing while maintaining computational efficiency. However, there are still considerations, as the performance of proximal guidance can be sensitive to hyperparameters. Exploring heuristics or automated methods for parameter selection could enhance the usability and generalizability of the proposed method. Our work demonstrates the potential of proximal guidance and opens avenues for further research in tuning-free real image editing.

{\small
\bibliographystyle{ieee_fullname}
\bibliography{example_paper}

\begin{thebibliography}{10}\itemsep=-1pt

\bibitem{abdal2022clip2stylegan}
Rameen Abdal, Peihao Zhu, John Femiani, Niloy Mitra, and Peter Wonka.
\newblock Clip2stylegan: Unsupervised extraction of stylegan edit directions.
\newblock In {\em ACM SIGGRAPH 2022 Conference Proceedings}, pages 1--9, 2022.

\bibitem{abdal2021styleflow}
Rameen Abdal, Peihao Zhu, Niloy~J Mitra, and Peter Wonka.
\newblock Styleflow: Attribute-conditioned exploration of stylegan-generated
  images using conditional continuous normalizing flows.
\newblock {\em ACM Transactions on Graphics (ToG)}, 40(3):1--21, 2021.

\bibitem{avrahami2023break}
Omri Avrahami, Kfir Aberman, Ohad Fried, Daniel Cohen-Or, and Dani Lischinski.
\newblock Break-a-scene: Extracting multiple concepts from a single image.
\newblock {\em arXiv preprint arXiv:2305.16311}, 2023.

\bibitem{boyd2011distributed}
Stephen Boyd, Neal Parikh, Eric Chu, Borja Peleato, Jonathan Eckstein, et~al.
\newblock Distributed optimization and statistical learning via the alternating
  direction method of multipliers.
\newblock {\em Foundations and Trends{\textregistered} in Machine learning},
  3(1):1--122, 2011.

\bibitem{brack2023sega}
Manuel Brack, Felix Friedrich, Dominik Hintersdorf, Lukas Struppek, Patrick
  Schramowski, and Kristian Kersting.
\newblock Sega: Instructing diffusion using semantic dimensions.
\newblock {\em arXiv preprint arXiv:2301.12247}, 2023.

\bibitem{cao2023masactrl}
Mingdeng Cao, Xintao Wang, Zhongang Qi, Ying Shan, Xiaohu Qie, and Yinqiang
  Zheng.
\newblock Masactrl: Tuning-free mutual self-attention control for consistent
  image synthesis and editing.
\newblock {\em arXiv preprint arXiv:2304.08465}, 2023.

\bibitem{chang2023muse}
Huiwen Chang, Han Zhang, Jarred Barber, AJ Maschinot, Jose Lezama, Lu Jiang,
  Ming-Hsuan Yang, Kevin Murphy, William~T Freeman, Michael Rubinstein, et~al.
\newblock Muse: Text-to-image generation via masked generative transformers.
\newblock {\em arXiv preprint arXiv:2301.00704}, 2023.

\bibitem{chefer2023hidden}
Hila Chefer, Oran Lang, Mor Geva, Volodymyr Polosukhin, Assaf Shocher, Michal
  Irani, Inbar Mosseri, and Lior Wolf.
\newblock The hidden language of diffusion models.
\newblock {\em arXiv preprint arXiv:2306.00966}, 2023.

\bibitem{chen2023disenbooth}
Hong Chen, Yipeng Zhang, Xin Wang, Xuguang Duan, Yuwei Zhou, and Wenwu Zhu.
\newblock Disenbooth: Identity-preserving disentangled tuning for
  subject-driven text-to-image generation, 2023.

\bibitem{chen2023revisiting}
Yuxiao Chen, Jianbo Yuan, Yu Tian, Shijie Geng, Xinyu Li, Ding Zhou,
  Dimitris~N. Metaxas, and Hongxia Yang.
\newblock Revisiting multimodal representation in contrastive learning: from
  patch and token embeddings to finite discrete tokens.
\newblock In {\em Proceedings of the IEEE/CVF Conference on Computer Vision and
  Pattern Recognition}, 2023.

\bibitem{daras2022multiresolution}
Giannis Daras and Alexandros~G Dimakis.
\newblock Multiresolution textual inversion.
\newblock {\em arXiv preprint arXiv:2211.17115}, 2022.

\bibitem{dutt2023parameter}
Raman Dutt, Linus Ericsson, Pedro Sanchez, Sotirios~A Tsaftaris, and Timothy
  Hospedales.
\newblock Parameter-efficient fine-tuning for medical image analysis: The
  missed opportunity.
\newblock {\em arXiv preprint arXiv:2305.08252}, 2023.

\bibitem{fessler2021prox}
Jeffrey~A. Fessler.
\newblock Lecture notes on proximal methods.
\newblock \url{https://web.eecs.umich.edu/~fessler/course/598/l/n-05-prox.pdf},
  2021.

\bibitem{gal2022image}
Rinon Gal, Yuval Alaluf, Yuval Atzmon, Or Patashnik, Amit~H Bermano, Gal
  Chechik, and Daniel Cohen-Or.
\newblock An image is worth one word: Personalizing text-to-image generation
  using textual inversion.
\newblock {\em arXiv preprint arXiv:2208.01618}, 2022.

\bibitem{hiclip}
Shijie Geng, Jianbo Yuan, Yu Tian, Yuxiao Chen, and Yongfeng Zhang.
\newblock Hiclip: Contrastive language-image pretraining with hierarchy-aware
  attention.
\newblock {\em arXiv preprint arXiv:2303.02995}, 2023.

\bibitem{gu2022vector}
Shuyang Gu, Dong Chen, Jianmin Bao, Fang Wen, Bo Zhang, Dongdong Chen, Lu Yuan,
  and Baining Guo.
\newblock Vector quantized diffusion model for text-to-image synthesis.
\newblock In {\em Proceedings of the IEEE/CVF Conference on Computer Vision and
  Pattern Recognition}, pages 10696--10706, 2022.

\bibitem{han2020robust}
Ligong Han, Ruijiang Gao, Mun Kim, Xin Tao, Bo Liu, and Dimitris Metaxas.
\newblock Robust conditional gan from uncertainty-aware pairwise comparisons.
\newblock In {\em Proceedings of the AAAI Conference on Artificial
  Intelligence}, volume~34, pages 10909--10916, 2020.

\bibitem{han2023svdiff}
Ligong Han, Yinxiao Li, Han Zhang, Peyman Milanfar, Dimitris Metaxas, and Feng
  Yang.
\newblock Svdiff: Compact parameter space for diffusion fine-tuning.
\newblock {\em arXiv preprint arXiv:2303.11305}, 2023.

\bibitem{han2021dual}
Ligong Han, Martin~Renqiang Min, Anastasis Stathopoulos, Yu Tian, Ruijiang Gao,
  Asim Kadav, and Dimitris~N Metaxas.
\newblock Dual projection generative adversarial networks for conditional image
  generation.
\newblock In {\em Proceedings of the IEEE/CVF International Conference on
  Computer Vision}, pages 14438--14447, 2021.

\bibitem{han2022ae}
Ligong Han, Sri~Harsha Musunuri, Martin~Renqiang Min, Ruijiang Gao, Yu Tian,
  and Dimitris Metaxas.
\newblock Ae-stylegan: Improved training of style-based auto-encoders.
\newblock In {\em Proceedings of the IEEE/CVF Winter Conference on Applications
  of Computer Vision}, pages 3134--3143, 2022.

\bibitem{han2022show}
Ligong Han, Jian Ren, Hsin-Ying Lee, Francesco Barbieri, Kyle Olszewski,
  Shervin Minaee, Dimitris Metaxas, and Sergey Tulyakov.
\newblock Show me what and tell me how: Video synthesis via multimodal
  conditioning.
\newblock In {\em Proceedings of the IEEE/CVF Conference on Computer Vision and
  Pattern Recognition}, pages 3615--3625, 2022.

\bibitem{han2020unbiased}
Ligong Han, Anastasis Stathopoulos, Tao Xue, and Dimitris Metaxas.
\newblock Unbiased auxiliary classifier gans with mine.
\newblock {\em arXiv preprint arXiv:2006.07567}, 2020.

\bibitem{harkonen2020ganspace}
Erik H{\"a}rk{\"o}nen, Aaron Hertzmann, Jaakko Lehtinen, and Sylvain Paris.
\newblock Ganspace: Discovering interpretable gan controls.
\newblock {\em Advances in Neural Information Processing Systems},
  33:9841--9850, 2020.

\bibitem{hertz2023delta}
Amir Hertz, Kfir Aberman, and Daniel Cohen-Or.
\newblock Delta denoising score.
\newblock {\em arXiv preprint arXiv:2304.07090}, 2023.

\bibitem{hertz2022prompt}
Amir Hertz, Ron Mokady, Jay Tenenbaum, Kfir Aberman, Yael Pritch, and Daniel
  Cohen-Or.
\newblock Prompt-to-prompt image editing with cross attention control.
\newblock {\em arXiv preprint arXiv:2208.01626}, 2022.

\bibitem{ho2020denoising}
Jonathan Ho, Ajay Jain, and Pieter Abbeel.
\newblock Denoising diffusion probabilistic models.
\newblock {\em Advances in Neural Information Processing Systems},
  33:6840--6851, 2020.

\bibitem{ho2021classifier}
Jonathan Ho and Tim Salimans.
\newblock Classifier-free diffusion guidance.
\newblock In {\em NeurIPS 2021 Workshop on Deep Generative Models and
  Downstream Applications}, 2021.

\bibitem{hu2021lora}
Edward~J Hu, Yelong Shen, Phillip Wallis, Zeyuan Allen-Zhu, Yuanzhi Li, Shean
  Wang, Lu Wang, and Weizhu Chen.
\newblock Lora: Low-rank adaptation of large language models.
\newblock {\em arXiv preprint arXiv:2106.09685}, 2021.

\bibitem{huang2023zero}
Yihao Huang, Qing Guo, and Felix Juefei-Xu.
\newblock Zero-day backdoor attack against text-to-image diffusion models via
  personalization.
\newblock {\em arXiv preprint arXiv:2305.10701}, 2023.

\bibitem{jiang2023avatarcraft}
Ruixiang Jiang, Can Wang, Jingbo Zhang, Menglei Chai, Mingming He, Dongdong
  Chen, and Jing Liao.
\newblock Avatarcraft: Transforming text into neural human avatars with
  parameterized shape and pose control.
\newblock {\em arXiv preprint arXiv:2303.17606}, 2023.

\bibitem{kumari2022multi}
Nupur Kumari, Bingliang Zhang, Richard Zhang, Eli Shechtman, and Jun-Yan Zhu.
\newblock Multi-concept customization of text-to-image diffusion.
\newblock {\em Proceedings of the IEEE/CVF Conference on Computer Vision and
  Pattern Recognition (CVPR)}, 2023.

\bibitem{lee2023soundini}
Seung~Hyun Lee, Sieun Kim, Innfarn Yoo, Feng Yang, Donghyeon Cho, Youngseo Kim,
  Huiwen Chang, Jinkyu Kim, and Sangpil Kim.
\newblock Soundini: Sound-guided diffusion for natural video editing.
\newblock {\em arXiv preprint arXiv:2304.06818}, 2023.

\bibitem{li2023layerdiffusion}
Pengzhi Li, QInxuan Huang, Yikang Ding, and Zhiheng Li.
\newblock Layerdiffusion: Layered controlled image editing with diffusion
  models.
\newblock {\em arXiv preprint arXiv:2305.18676}, 2023.

\bibitem{li2023stylediffusion}
Senmao Li, Joost van~de Weijer, Taihang Hu, Fahad~Shahbaz Khan, Qibin Hou,
  Yaxing Wang, and Jian Yang.
\newblock Stylediffusion: Prompt-embedding inversion for text-based editing.
\newblock {\em arXiv preprint arXiv:2303.15649}, 2023.

\bibitem{li2023gligen}
Yuheng Li, Haotian Liu, Qingyang Wu, Fangzhou Mu, Jianwei Yang, Jianfeng Gao,
  Chunyuan Li, and Yong~Jae Lee.
\newblock Gligen: Open-set grounded text-to-image generation.
\newblock {\em arXiv preprint arXiv:2301.07093}, 2023.

\bibitem{li2023snapfusion}
Yanyu Li, Huan Wang, Qing Jin, Ju Hu, Pavlo Chemerys, Yun Fu, Yanzhi Wang,
  Sergey Tulyakov, and Jian Ren.
\newblock Snapfusion: Text-to-image diffusion model on mobile devices within
  two seconds.
\newblock {\em arXiv preprint arXiv:2306.00980}, 2023.

\bibitem{miyake2023negative}
Daiki Miyake, Akihiro Iohara, Yu Saito, and Toshiyuki Tanaka.
\newblock Negative-prompt inversion: Fast image inversion for editing with
  text-guided diffusion models.
\newblock {\em arXiv preprint arXiv:2305.16807}, 2023.

\bibitem{mokady2022null}
Ron Mokady, Amir Hertz, Kfir Aberman, Yael Pritch, and Daniel Cohen-Or.
\newblock Null-text inversion for editing real images using guided diffusion
  models.
\newblock {\em arXiv preprint arXiv:2211.09794}, 2022.

\bibitem{mou2023t2i}
Chong Mou, Xintao Wang, Liangbin Xie, Jian Zhang, Zhongang Qi, Ying Shan, and
  Xiaohu Qie.
\newblock T2i-adapter: Learning adapters to dig out more controllable ability
  for text-to-image diffusion models.
\newblock {\em arXiv preprint arXiv:2302.08453}, 2023.

\bibitem{nichol2021glide}
Alex Nichol, Prafulla Dhariwal, Aditya Ramesh, Pranav Shyam, Pamela Mishkin,
  Bob McGrew, Ilya Sutskever, and Mark Chen.
\newblock Glide: Towards photorealistic image generation and editing with
  text-guided diffusion models.
\newblock {\em arXiv preprint arXiv:2112.10741}, 2021.

\bibitem{nichol2021improved}
Alexander~Quinn Nichol and Prafulla Dhariwal.
\newblock Improved denoising diffusion probabilistic models.
\newblock In {\em International Conference on Machine Learning}, pages
  8162--8171. PMLR, 2021.

\bibitem{oh2001image}
Byong~Mok Oh, Max Chen, Julie Dorsey, and Fr{\'e}do Durand.
\newblock Image-based modeling and photo editing.
\newblock In {\em Proceedings of the 28th annual conference on Computer
  graphics and interactive techniques}, pages 433--442, 2001.

\bibitem{orgad2023editing}
Hadas Orgad, Bahjat Kawar, and Yonatan Belinkov.
\newblock Editing implicit assumptions in text-to-image diffusion models.
\newblock {\em arXiv preprint arXiv:2303.08084}, 2023.

\bibitem{patashnik2021styleclip}
Or Patashnik, Zongze Wu, Eli Shechtman, Daniel Cohen-Or, and Dani Lischinski.
\newblock Styleclip: Text-driven manipulation of stylegan imagery.
\newblock {\em arXiv preprint arXiv:2103.17249}, 2021.

\bibitem{qi2023fatezero}
Chenyang Qi, Xiaodong Cun, Yong Zhang, Chenyang Lei, Xintao Wang, Ying Shan,
  and Qifeng Chen.
\newblock Fatezero: Fusing attentions for zero-shot text-based video editing.
\newblock {\em arXiv preprint arXiv:2303.09535}, 2023.

\bibitem{ramesh2022hierarchical}
Aditya Ramesh, Prafulla Dhariwal, Alex Nichol, Casey Chu, and Mark Chen.
\newblock Hierarchical text-conditional image generation with clip latents.
\newblock {\em arXiv preprint arXiv:2204.06125}, 2022.

\bibitem{ramesh2021zero}
Aditya Ramesh, Mikhail Pavlov, Gabriel Goh, Scott Gray, Chelsea Voss, Alec
  Radford, Mark Chen, and Ilya Sutskever.
\newblock Zero-shot text-to-image generation.
\newblock {\em arXiv preprint arXiv:2102.12092}, 2021.

\bibitem{ren2022image}
Mengwei Ren, Mauricio Delbracio, Hossein Talebi, Guido Gerig, and Peyman
  Milanfar.
\newblock Image deblurring with domain generalizable diffusion models.
\newblock {\em arXiv preprint arXiv:2212.01789}, 2022.

\bibitem{rombach2022high}
Robin Rombach, Andreas Blattmann, Dominik Lorenz, Patrick Esser, and Bj\"orn
  Ommer.
\newblock High-resolution image synthesis with latent diffusion models.
\newblock In {\em Proceedings of the IEEE/CVF Conference on Computer Vision and
  Pattern Recognition (CVPR)}, pages 10684--10695, June 2022.

\bibitem{ruiz2022dreambooth}
Nataniel Ruiz, Yuanzhen Li, Varun Jampani, Yael Pritch, Michael Rubinstein, and
  Kfir Aberman.
\newblock Dreambooth: Fine tuning text-to-image diffusion models for
  subject-driven generation.
\newblock {\em arXiv preprint arXiv:2208.12242}, 2022.

\bibitem{saharia2022photorealistic}
Chitwan Saharia, William Chan, Saurabh Saxena, Lala Li, Jay Whang, Emily
  Denton, Seyed Kamyar~Seyed Ghasemipour, Burcu~Karagol Ayan, S~Sara Mahdavi,
  Rapha~Gontijo Lopes, et~al.
\newblock Photorealistic text-to-image diffusion models with deep language
  understanding.
\newblock {\em arXiv preprint arXiv:2205.11487}, 2022.

\bibitem{shi2023instantbooth}
Jing Shi, Wei Xiong, Zhe Lin, and Hyun~Joon Jung.
\newblock Instantbooth: Personalized text-to-image generation without test-time
  finetuning.
\newblock {\em arXiv preprint arXiv:2304.03411}, 2023.

\bibitem{sohl2015deep}
Jascha Sohl-Dickstein, Eric Weiss, Niru Maheswaranathan, and Surya Ganguli.
\newblock Deep unsupervised learning using nonequilibrium thermodynamics.
\newblock In {\em International Conference on Machine Learning}, pages
  2256--2265. PMLR, 2015.

\bibitem{sohn2023styledrop}
Kihyuk Sohn, Nataniel Ruiz, Kimin Lee, Daniel~Castro Chin, Irina Blok, Huiwen
  Chang, Jarred Barber, Lu Jiang, Glenn Entis, Yuanzhen Li, et~al.
\newblock Styledrop: Text-to-image generation in any style.
\newblock {\em arXiv preprint arXiv:2306.00983}, 2023.

\bibitem{song2021denoising}
Jiaming Song, Chenlin Meng, and Stefano Ermon.
\newblock Denoising diffusion implicit models.
\newblock In {\em International Conference on Learning Representations}, 2021.

\bibitem{song2023consistency}
Yang Song, Prafulla Dhariwal, Mark Chen, and Ilya Sutskever.
\newblock Consistency models.
\newblock {\em arXiv preprint arXiv:2303.01469}, 2023.

\bibitem{song2019generative}
Yang Song and Stefano Ermon.
\newblock Generative modeling by estimating gradients of the data distribution.
\newblock {\em Advances in Neural Information Processing Systems}, 32, 2019.

\bibitem{song2020score}
Yang Song, Jascha Sohl-Dickstein, Diederik~P Kingma, Abhishek Kumar, Stefano
  Ermon, and Ben Poole.
\newblock Score-based generative modeling through stochastic differential
  equations.
\newblock {\em arXiv preprint arXiv:2011.13456}, 2020.

\bibitem{tao2020df}
Ming Tao, Hao Tang, Songsong Wu, Nicu Sebe, Xiao-Yuan Jing, Fei Wu, and Bingkun
  Bao.
\newblock Df-gan: Deep fusion generative adversarial networks for text-to-image
  synthesis.
\newblock {\em arXiv preprint arXiv:2008.05865}, 2020.

\bibitem{tibshirani2019prox}
Ryan Tibshirani.
\newblock Lecture notes on proximal gradient descent.
\newblock
  \url{https://www.stat.cmu.edu/~ryantibs/convexopt/lectures/prox-grad.pdf},
  2019.

\bibitem{tumanyan2023plug}
Narek Tumanyan, Michal Geyer, Shai Bagon, and Tali Dekel.
\newblock Plug-and-play diffusion features for text-driven image-to-image
  translation.
\newblock In {\em Proceedings of the IEEE/CVF Conference on Computer Vision and
  Pattern Recognition}, pages 1921--1930, 2023.

\bibitem{voynov2023p+}
Andrey Voynov, Qinghao Chu, Daniel Cohen-Or, and Kfir Aberman.
\newblock $ p+ $: Extended textual conditioning in text-to-image generation.
\newblock {\em arXiv preprint arXiv:2303.09522}, 2023.

\bibitem{wang2023compositional}
Ruichen Wang, Zekang Chen, Chen Chen, Jian Ma, Haonan Lu, and Xiaodong Lin.
\newblock Compositional text-to-image synthesis with attention map control of
  diffusion models.
\newblock {\em arXiv preprint arXiv:2305.13921}, 2023.

\bibitem{wang2023patch}
Zhendong Wang, Yifan Jiang, Huangjie Zheng, Peihao Wang, Pengcheng He,
  Zhangyang Wang, Weizhu Chen, and Mingyuan Zhou.
\newblock Patch diffusion: Faster and more data-efficient training of diffusion
  models.
\newblock {\em arXiv preprint arXiv:2304.12526}, 2023.

\bibitem{wu2023sin3dm}
Rundi Wu, Ruoshi Liu, Carl Vondrick, and Changxi Zheng.
\newblock Sin3dm: Learning a diffusion model from a single 3d textured shape.
\newblock {\em arXiv preprint arXiv:2305.15399}, 2023.

\bibitem{xu2018attngan}
Tao Xu, Pengchuan Zhang, Qiuyuan Huang, Han Zhang, Zhe Gan, Xiaolei Huang, and
  Xiaodong He.
\newblock Attngan: Fine-grained text to image generation with attentional
  generative adversarial networks.
\newblock In {\em Proceedings of the IEEE conference on computer vision and
  pattern recognition}, pages 1316--1324, 2018.

\bibitem{ye2021improving}
Hui Ye, Xiulong Yang, Martin Takac, Rajshekhar Sunderraman, and Shihao Ji.
\newblock Improving text-to-image synthesis using contrastive learning.
\newblock {\em arXiv preprint arXiv:2107.02423}, 2021.

\bibitem{yu2022scaling}
Jiahui Yu, Yuanzhong Xu, Jing~Yu Koh, Thang Luong, Gunjan Baid, Zirui Wang,
  Vijay Vasudevan, Alexander Ku, Yinfei Yang, Burcu~Karagol Ayan, et~al.
\newblock Scaling autoregressive models for content-rich text-to-image
  generation.
\newblock {\em arXiv preprint arXiv:2206.10789}, 2022.

\bibitem{zhan2021multimodal}
Fangneng Zhan, Yingchen Yu, Rongliang Wu, Jiahui Zhang, Shijian Lu, Lingjie
  Liu, Adam Kortylewski, Christian Theobalt, and Eric Xing.
\newblock Multimodal image synthesis and editing: A survey.
\newblock {\em arXiv preprint arXiv:2112.13592}, 2021.

\bibitem{zhang2021cross}
Han Zhang, Jing~Yu Koh, Jason Baldridge, Honglak Lee, and Yinfei Yang.
\newblock Cross-modal contrastive learning for text-to-image generation.
\newblock In {\em Proceedings of the IEEE/CVF conference on computer vision and
  pattern recognition}, pages 833--842, 2021.

\bibitem{zhang2023robustness}
Jiaxin Zhang, Kamalika Das, and Sricharan Kumar.
\newblock On the robustness of diffusion inversion in image manipulation.
\newblock In {\em ICLR 2023 Workshop on Trustworthy and Reliable Large-Scale
  Machine Learning Models}, 2023.

\bibitem{zhang2023prospect}
Yuxin Zhang, Weiming Dong, Fan Tang, Nisha Huang, Haibin Huang, Chongyang Ma,
  Tong-Yee Lee, Oliver Deussen, and Changsheng Xu.
\newblock Prospect: Expanded conditioning for the personalization of
  attribute-aware image generation.
\newblock {\em arXiv preprint arXiv:2305.16225}, 2023.

\bibitem{zhang2023real}
Yuechen Zhang, Jinbo Xing, Eric Lo, and Jiaya Jia.
\newblock Real-world image variation by aligning diffusion inversion chain.
\newblock {\em arXiv preprint arXiv:2305.18729}, 2023.

\bibitem{zhang2023text}
Zhongping Zhang, Jian Zheng, Jacob~Zhiyuan Fang, and Bryan~A Plummer.
\newblock Text-to-image editing by image information removal.
\newblock {\em arXiv preprint arXiv:2305.17489}, 2023.

\bibitem{zhu2019dm}
Minfeng Zhu, Pingbo Pan, Wei Chen, and Yi Yang.
\newblock Dm-gan: Dynamic memory generative adversarial networks for
  text-to-image synthesis.
\newblock In {\em Proceedings of the IEEE/CVF Conference on Computer Vision and
  Pattern Recognition}, pages 5802--5810, 2019.

\end{thebibliography}
}

\newpage
\appendix
\onecolumn

\section{Proof of Remark~\ref{remark:1}}
\begin{remark}
    Negative-prompt inversion is the {\em exact} closed-form solution if we solve null-text inversion optimizations to track the DDIM reconstruction trajectory $\{\hat{z}_t\}$, with $\bar{z}_T$ initialized as $\hat{z}_T=z^*_T$,
    \begin{align}
        C=\argmin_{\emptyset_t}||z_{t-1}(\bar{z_t}, \emptyset_t, C)-\hat{z}_{t-1}||_2^2.
    \end{align}
\end{remark}
\begin{proof}
    Following negative-prompt inversion~\cite{miyake2023negative}, we prove this by induction. Suppose at timestep $t$, $\emptyset_t=C$ and $\bar{z}_t=\hat{z}_t$ hold, then we derive $\bar{z}_{t-1}$ for timestep $t-1$.
    By definition (\cref{eq:ddim_reverse} with classifier-free guidance),
    \begin{align}
        \bar{z}_{t-1}=z_{t-1}(\bar{z}_{t},t,C,\emptyset_t)=\frac{\sqrt{\alpha_{t-1}}}{{\sqrt\alpha_{t}}}\bar{z}_t + \sqrt{\alpha_{t-1}} \left(\sqrt{\frac{1}{\alpha_{t-1}}-1} - \sqrt{\frac{1}{\alpha_{t}}-1}\right)\tilde{\epsilon}_{\theta}(\bar{z}_t, t, C, \emptyset_t).\label{eq:npi1}
    \end{align}
    Since $\bar{z}_t=\hat{z}_t$ and by \cref{eq:ddim_forward_exact},
    \begin{align}
        \bar{z}_t=\hat{z}_{t}=\frac{\sqrt{\alpha_{t}}}{{\sqrt{\alpha_{t-1}}}}\hat{z}_{t-1} + \sqrt{\alpha_{t}} \left(\sqrt{\frac{1}{\alpha_{t}}-1} - \sqrt{\frac{1}{\alpha_{t-1}}-1}\right)\epsilon_{\theta}(\hat{z}_{t}, t, C).
    \end{align}
    Substituting the above into \cref{eq:npi1}, we have
    \begin{align}
        \bar{z}_{t-1}=\hat{z}_{t-1} + \sqrt{\alpha_{t-1}} \left(\sqrt{\frac{1}{\alpha_{t-1}}-1} - \sqrt{\frac{1}{\alpha_{t}}-1}\right)\left(\tilde{\epsilon}_{\theta}(\hat{z}_t, t, C, \emptyset_t) - \epsilon_{\theta}(\hat{z}_{t}, t, C)\right).
    \end{align}
    \noindent Since $\tilde{\epsilon}_{\theta}(\hat{z}_t, t, C, \emptyset_t)-\epsilon_\theta(\hat{z}_t, t, C)=(w-1)\left( \epsilon_\theta(\hat{z}_t, t, C) - \epsilon_\theta(\hat{z}_t, t, \emptyset_t)\right)$, we have $\bar{z}_{t-1}=\hat{z}_{t-1}$ if $\emptyset_t=C$.
\end{proof}

\section{Reconstruction Guidance}
\begin{figure*}[t]
  \centering
  \includegraphics[width=0.95\linewidth]{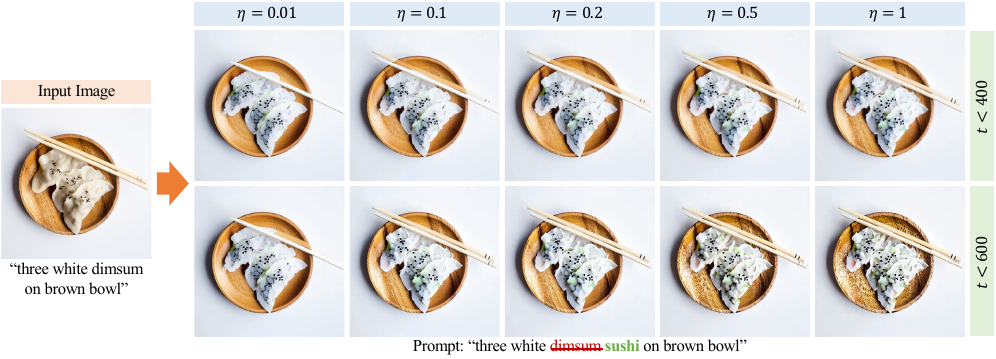}
\caption{\textbf{Ablation study of reconstruction guidance.} The figure shows visual results obtained by varying the stepsize of performing reconstruction guidance $\eta$ from the 0.01 to 1. The first row represents performing guidance when $t<400$, while the second shows the effects of $t<600$. The threshold is set to the 70\% quantile and hard-thresholding is used.}
  \label{fig:ablate_recon}
\end{figure*}

\begin{algorithm}[h]
	\renewcommand{\algorithmicrequire}{\textbf{Input:}}
	\renewcommand{\algorithmicensure}{\textbf{Output:}}
	\caption{Proximal Negative-Prompt Inversion with reconstruction guidance}
	\label{alg:2}
	\begin{algorithmic}[1]
		\REQUIRE Given source original sample $z_0$, source condition $C$, target condition $C'$, denoising model $\epsilon_{\theta}$, proximal function $\text{prox}_{\lambda}(\cdot)$.
        \STATE $\bar {z}_T=\text{DDIMInvert}(z_0, C, w=1)$
            \STATE $\tilde z_T = {\bar {z}_T}$
		\FOR{$t = T$ to $1$}
            \STATE $\tilde\epsilon_{src} = \epsilon_{\theta}(\tilde z_t, t, C)$
            \STATE $\tilde\epsilon_{tar} = \epsilon_{\theta}(\tilde z_t, t, C')$
            \STATE $\tilde\epsilon = {\tilde\epsilon}_{src} + w \cdot \text{prox}_{\lambda}(\tilde\epsilon_{tar}- \tilde\epsilon_{src})$
            \STATE $M = |\tilde\epsilon_{tar}- \tilde\epsilon_{src}| \leq \lambda$
            \STATE $\tilde z_0 = \frac{1}{\sqrt{\alpha_t}}\tilde z_t - \sqrt{\frac{1}{\alpha_t}-1}\tilde\epsilon$
            \IF {reconstruction guidance \AND $t<T_{rec}$}
                \STATE $\tilde z_0 = \tilde z_0 - \eta M\odot(\tilde z_0 - z_0)$
            \ENDIF
            \STATE $\tilde z_{t-1} = \sqrt{\alpha_{t-1}}\tilde z_0+\sqrt{1-\alpha_{t-1}}\tilde\epsilon$
            \IF {inversion guidance \AND $t<T_{inv}$}
                \STATE $\tilde z_{t-1} = \tilde z_{t-1} - \eta M\odot(\tilde z_{t-1} - z^*_{t-1})$
            \ENDIF
		\ENDFOR
		\STATE \textbf{return} $\tilde z_0$
	\end{algorithmic}  
\end{algorithm}
We have introduced the concept of ``reconstruction guidance'' as an additional solution to address the issue of imperfect DDIM reconstruction. Another straight-forward solution is {\em reconstruction guidance}. To do so, we perform one step of gradient descent on the current predicted original sample $\tilde z_0$ to align it with the source sample $z_0$. Similarly, this gradient descent step is applied to the ``unedited'' region identified by the mask $M = |\tilde\epsilon_{tar} - \tilde\epsilon_{src}| \leq \lambda$. The update can be expressed as $\tilde z_0 \leftarrow \tilde z_0 - \eta M \odot (\tilde z_0 - z_0)$. The algorithm with reconstruction guidance is outlined in \cref{alg:2}. In \cref{fig:ablate_recon}, we present visual results obtained by varying the stepsize $\eta$. The guidance is applied when $t<T_{rec}$. As observed, when the guidance strength is small (with a small $\eta$), the reconstruction of chopsticks is incomplete. Increasing $T_{rec}$ results in accurate reconstruction of the chopsticks, however, a large $\eta$ may introduce artifacts such as an over-amplified contrast ratio. Based on empirical findings, we generally set $T_{rec}=400$ and $\eta=0.1$, although inversion guidance is still preferred over reconstruction guidance.

\end{document}